# Toward Integrated Human-machine Intelligence for Civil Engineering: An Interdisciplinary Perspective

Cheng Zhang, Ph.D., S.M.ASCE,[1] Jinwoo Kim, S.M.ASCE,[2] JungHo Jeon,[3] Jinding Xing,[4] Changbum Ahn, Ph.D., A.M.ASCE,[5] Pingbo Tang, Ph.D., PE, M.ASCE,[6] and Hubo Cai, Ph.D., PE, M.ASCE[7]

[1]Department of Construction Science and Organizational Leadership, Purdue University Northwest, 2200 169th Street, Hammond, IN, 46323; email: zhan4168@pnw.edu
[2]Department of Multidisciplinary Engineering, Texas A&M University, 3137 TAMU, College Station, TX 77840; email: jwkim@tamu.edu
[3]School of Civil Engineering, Purdue University, 550 Stadium Mall Drive, West Lafayette, IN 47907-2051; email: jeon77@purdue.edu
[4]Department of Civil and Environmental Engineering, Carnegie Mellon University, Porter Hall 119 Pittsburgh, PA 15213-3890; email: jindingx@andrew.cmu.edu
[5]Department of Construction Science, Texas A&M University, 3137 TAMU, College Station, TX 77840; email: ryanahn@tamu.edu
[6]Department of Civil and Environmental Engineering, Carnegie Mellon University, Porter Hall 119 Pittsburgh, PA 15213-3890; email: ptang@andrew.cmu.edu
[7]School of Civil Engineering, Purdue University, 550 Stadium Mall Drive, West Lafayette, IN 47907-2051; email: hubocai@purdue.edu

**ABSTRACT**

The purpose of this paper is to examine the opportunities and barriers of Integrated Human-Machine Intelligence (IHMI) in civil engineering. Integrating artificial intelligence's high efficiency and repeatability with humans' adaptability in various contexts can advance timely and reliable decision-making during civil engineering projects and emergencies. Successful cases in other domains, such as biomedical science, healthcare, and transportation, showed the potential of IHMI in data-driven, knowledge-based decision-making in numerous civil engineering applications. However, whether the industry and academia are ready to embrace the era of IHMI and maximize its benefit to the industry is still questionable due to several knowledge gaps. This paper thus calls for future studies in exploring the value, method, and challenges of applying IHMI in civil engineering. Our systematic review of the literature and motivating cases has identified four knowledge gaps in achieving effective IHMI in civil engineering. First, it is unknown what types of tasks in the civil engineering domain can be assisted by AI and to what extent. Second, the interface between human and AI in civil engineering-related tasks need more precise and formal definition. Third, the barriers that impede collecting detailed behavioral data from humans and contextual environments deserve systematic classification and prototyping. Lastly, it is unknown what expected and unexpected impacts will IHMI have on the AEC industry and entrepreneurship. Analyzing these knowledge gaps led to a list of identified research questions. This paper will lay the foundation for identifying relevant studies to form a research roadmap to address the four knowledge gaps identified.

# INTRODUCTION

In recent years, the civil engineering domain experienced significant development in harnessing the power of artificial intelligence. Artificial intelligence (AI), also referred to as machine intelligence, is a powerful tool to address the civil engineering domain problems that traditional computational approaches cannot address. It can benefit the civil engineering domain by providing automation, digitization, and data fusion capacity, with high accuracy and efficiency. Existing review papers highlighted AI's contribution in advancing structural health monitoring, infrastructure sustainability analysis, optimization in structural design, construction safety monitoring, disaster response, and more (Pan and Zhang 2021).

Despite the breakthroughs of AI-based approaches, some practical challenges in the civil engineering domain impede the effective application of the current AI approaches. Figure 1 overviews the civil engineering domain's challenges and limitations of various AI approaches to address them. Overall, state-of-the-art AI methods have limitations in mitigating the challenges brought by the unique characteristics of civil engineering tasks: uniqueness, labor-intensive, dynamics, and uncertainty. Civil engineering projects' unique nature is rooted in the differences in project size, needs, methods, environment, and the vast stakeholders. Many civil engineering applications are manual and labor-intensive due to the highly uncertain natural and built civil engineering project environments. Such complexity and uncertainty cause unique challenges in harnessing AI's power by training neural network models using large datasets. First, the dynamic environment and human behavior will lead to noisy data. Furthermore, human-factor-related data may be unavailable due to privacy issues. Second, much domain knowledge lies in professionals' brains, and it is challenging to integrate such knowledge into AI models. Third, current AI models have poor explainability, and people lack trust in the results generated by AI models. Finally, most people working in the civil engineering domain have little or no coding experience, which causes difficulties in implementing and spreading AI-based approaches.

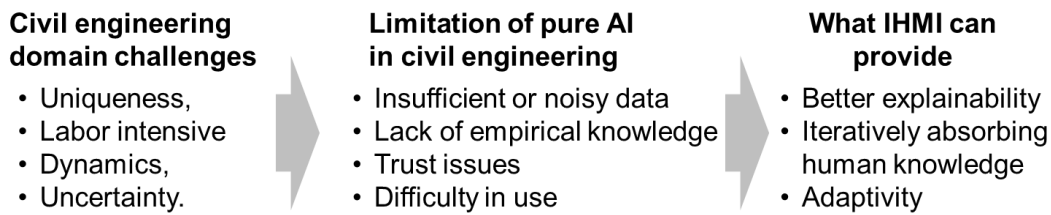

Figure 1. Overview of civil engineering domain challenges, limitation of pure AI approaches, and the benefit of IHMI.

Integrated human-machine intelligence (IHMI), also called human-AI integration, refers to the decision-making approaches that integrate humans' cognitive thinking and reasoning ability and the data mining and knowledge discovery ability of computers (National Science Foundation, n.d.). IHMI approaches can evolve with the addition of new knowledge and data. The right part of Figure 1 summarizes the potential benefit of IMHI in the civil engineering domain. However, how the industry and academia could embrace the era of IHMI is unknown. The civil engineering community lacks an in-depth understanding and systematic classification of technical challenges, practical issues (social value, ethical considerations, privacy, policy issues, business model considerations), and knowledge gaps. Thus, this paper calls for studies in characterizing human-AI integration in civil engineering through classifying practical domain tasks, core AI technology suitable for various tasks, human factors, and entrepreneurship issues. Such a characterization then reveals a list of research questions to bridge knowledge gaps on the four aspects of such classification efforts.

**INTEGRATED HUMAN-MACHINE INTELLIGENCE**

IHMI can potentially capture insights in addressing the challenges caused by civil engineering tasks in complex and dynamic contexts for several reasons (National Science Foundation, n.d.). First, intuitive common sense and domain knowledge embedded in the human brain are crucial to boosting machine learning algorithms' performance. Second, human experience can help select important parameters and hyperparameters of AI models, significantly increasing computational efficiency. Third, humans can extract useful information from noisy data, which helps mitigate missing, faulty, or low-resolution data in developing AI models. Finally, most AI models have poor explainability, but domain experts' interpretation can increase the trust in the results produced by AI. Based on these benefits, IHMI has great potential to boost various civil engineering tasks' reliability and efficiency despite the challenges - uniqueness, labor-intensive dynamics, and contextual uncertainty.

Although the studies of IHMI in civil engineering are still in their infancy, existing approaches in other science and engineering areas have provided examples illustrating its success and necessity. In the medical domain, the quality of collected data sets, the generalizability of training sets, and systems validation are often subject to algorithm developers' discretion. As a result, the developed AI models' translatability and safety to other clinical circumstances are often unclear (Rawson et al. 2019). Furthermore, in a recent study on the public confidence in AI-driven decisions, citizens responded that although AI's decision-making had potential advantages in improving the accuracy of diagnosis, interaction and oversight from a medical professional remained significant (Rawson et al. 2019). These demonstrate the importance of IHMI application in the biomedical and healthcare domain. Researchers have focused on the importance of IHMI since solely relying on AI in an autonomous car could result in fatal accidents. AI can assist drivers to develop better risk awareness and engage in safe driving actions by supporting their decision-making process (Bellet et al. 2019). However, the sole reliance on AI is far from meeting complex operation requirements in a dynamic environment. Many researchers have explored external interfaces (e.g., displays), design components, and interactions between drivers and vehicles to comprehend AI's effectiveness in assisting drivers (Rouchitsas and Alm 2019).

**MOTIVATING CASES: OPPORTUNITIES AND CHALLENGES OF IHMI IN CIVIL ENGINEERING**

This subsection uses two motivating cases to illustrate the potential of IHMI in addressing the domain challenges. Specifically, the following sections discuss how IHMI could surpass conventional AI approaches in handling insufficient or noisy data, harvesting empirical knowledge from experienced professionals, and improving human trust in AI models' decision-making results.

*Construction safety management*

The recent advent of sensing technologies (e.g., wearable sensors, vision sensors) generated various AI applications in identifying or measuring workers' unsafe behaviors, including improper use of personal protective equipment, access to hazardous areas, and operating equipment against the procedures (Wang et al. 2019). However, identifying at-risk workers and establishing interventions are delicate and complicated tasks. The involvement of human intelligence in the process is critical due to the following reasons: (1) Several key safety-related indicators, including personal characteristics (e.g., personal traits) and safety attitudes, cannot be assessed solely based on sensory data and thus require human intelligence (e.g., safety managers' evaluation); and (2) disseminating the assessment results of machine intelligence to workers would cause trust issues due to poor interpretability and accountability of AI models. Lack of trustworthy AI would potentially undermine the effect of interventions to improve construction safety.

Two different approaches in developing IHMI applications to help address safety management tasks are envisioned to this end. One approach is to have human safety managers as final decision-makers while feeding MI-based measurement data to support their decisions. In this approach, AI-based measurement's explainability will be essential in facilitating human managers' interpretation of AI outputs based on field observations, while advancing explainability in AI models is still immature. Alternatively, human managers can feed input datasets (e.g., observation-based evaluation) into AI models. Then AI models will identify at-risk workers by leveraging sensor-based data and managers' field observations. However, generating such datasets for training AI models is challenging because having experienced safety engineers annotate many images, video clips, and tests requires excessive time and effort.

*Nuclear power plant facility management*

Most Nuclear Power Plants (NPPs) have advanced anomaly detection systems equipped to detect faults in a single component/sensor or simultaneous faults in multiple components/sensors. These anomaly detection systems adopt model-based and data-driven approaches capable of identifying minor and slowly evolving anomalies. In current practice, anomaly detection systems with built-in NPP models adopt advanced machine learning algorithms to learn abnormal patterns from large amounts of sensor data. Such anomaly detection systems rely on experienced operators to investigate the anomalies' root causes and generate recovery plans and prevention strategies. The firm reliance between human operators and anomaly detection systems will increase the human operator's mental workload, making it difficult to avoid human errors in the anomaly detection process involving many data sources that rapidly change (Sheridan 1981). Furthermore, the knowledge and experiences generated in the human diagnosis process could hardly be recorded and reused by the NPP workers.

One promising approach to boost better IHMI in the NPP anomaly detection process is to extract features or signatures from human diagnosis process data and built the extracted knowledge to NPP anomaly detection systems. Such knowledge-augmented and data-driven approaches enable the anomaly detection system to identify the root causes of the anomalies and recommend multiple recovery plans and prevention strategies (Yan and Yu 2019). The human operators will then be released from laborious observing and control activities across many changing data sources. In that case, humans will be only responsible for choosing the proper control actions recommended by AI based on their intuitive common sense and contextual knowledge not captured by machine learning algorithms. However, challenges exist in implementing IHMI for Nuclear power plant facility management. The complex interactions between NPP systems and the uncertainty of NPP components make it challenging to record the human operators' decision-making process and train AI models using such data. For example, it is unclear how to feed the secondary data (e.g., operating memos) that support experienced operators' decision-making to the current AI models in anomaly detection systems.

**KNOWLEDGE GAPS AND RESEARCH QUESTIONS**

The motivating cases show the promising future of IHMI in addressing civil engineering domain problems. However, the exploration of IHMI in the civil engineering domain is still in the early stage. This study identified knowledge gaps (KGs) in developing practical IHMI applications in civil engineering via content analysis of literature and panel discussion. A focused literature review was conducted to explore the knowledge gaps behind those practical challenges. Then, a panel discussion was held to summarize the identified KGs and related research questions; the panel was comprised of seven researchers who have 3-15 years of experience in developing AI applications in civil engineering. The following subsections introduce those identified KGs and research questions.

*KG 1: The characteristics of the civil engineering tasks that are suitable to integrate AI and to what extent*

Tasks in civil engineering are complex, often include the interactions of multiple entities (e.g., human, infrastructure, and environment), and might involve the cooperation of multiple stakeholders who have different roles, e.g., owners, engineers, and researchers. Decision-making is a critical step that directly determines the success of a task. AI tools help humans (e.g., contractors, construction safety engineers, and facility engineers) make optimal decisions (Wang et al. 2019; Yan and Yu 2019). While the preliminary studies have illustrated the usefulness of AI in civil engineering, very little attention has been paid to the configuration, balance, and trust between AI and humans in sharing the decision-making responsibility in dealing with tasks of varying complexity. Consequently, it is unclear which tasks in the civil engineering domain can be assisted/performed by AI and what role is appropriate for AI to assume – a massive hurdle towards maximizing AI's value in civil engineering. Therefore, there is a critical need to characterize civil engineering tasks in AI adaptability and configure the collaboration in decision-making between AI and humans. This KG formulates the following research questions:
- What are the characteristics of the tasks that have benefited from AI?
- What are the tasks that may not be possible to benefit from AI?
- How can we utilize developed AIs to make successful decisions on complex tasks?

*KG 2: The interface between humans and AI in civil engineering tasks*

AI techniques and algorithms have been shining in research topics such as architectural design, energy efficiency, reliability analysis in architecture, engineering, and construction (AEC). AI methods have the advantage of honing the efficiency and effectiveness of AEC tasks. Despite the broad impact of AI, AI can be simplified as a system capable of generating simple responses based on the input data with a significant amount of data labeled by humans. The challenge is that the interface that allows a machine to learn from humans automatically and adapt to human users' behavior is still in its infancy (Miraz, Ali, and Excell 2021). As a result, human experts or AI system users often need to rely on data-driven approaches to adjust AI systems to support various human users effectively. Such data-driven approaches include manual labeling of the data or recording users' latent metadata and decision-making processes while using the AI system. These human users could be: 1) experts that label and interpret the data for training the machine to complete similar data interpretation; 2) AI system users who could interact with the computer for searching and finding the results. Both types of interactions have significant needs for capturing and harnessing humans' cognitive thinking when using AI systems. It is unclear what human-AI interaction interfaces or communication protocols could support harnessing the experienced operators' decision-making process and train the AI models in anomaly detection systems. To bridge this KG, the following research questions are proposed:
- Who is using and/or will use AI in the civil engineering domain?
- What interface should AI in civil engineering use to communicate with professionals (e.g., workers, managers, engineers, and other professionals)?
- What verbal and nonverbal communication protocols between humans and machines fit the collaborative situation?
- Which safety issues may cause or mitigate human-AI integration in the civil engineering domain? How can IHMI impact the psychological health of human co-workers?
- How can we ensure transparency and explainability of machines' decision preparation processes to match human decision makers' cognitive capacity?

*KG 3: The barriers impeding training data collection from humans and environments*

While data collection to build AI applications is not easy in the civil engineering domain, data collection for integrated human-machine intelligence (IHMI) applications in the civil engineering domain poses particular challenges due to the type and complexity of problems to be tackled by IHMI. Complex problems and high-level decision tasks usually handle secondary data (historical project data such as daily work reports, production rates, work change orders) generated from project stakeholders (e.g., designers, subcontractors, construction managers, workers) rather than raw sensory data. Thus, the availability of data is limited. Furthermore, collecting such data would pose confidentiality and privacy issues, as those data contain information related to data providers' interests and intellectual properties (e.g., private parties) (Assaad et al. 2021).

In addition, data annotation in IHMI applications often requires significant domain-specific knowledge and collective decisions of multiple experts. However, relatively low-cost labeling methods (e.g., using existing labels, crowd-based labels) may not be applicable (Gondia et al. 2020). For example, the IHMI applications for construction safety violation detection would need ground-truth annotation for unsafe scenes under each of the given scenarios. This annotation will require the consolidation of multiple experts' judgments on each scenario (Wang et al. 2019). Overall, considering the challenges of data collection for IHMI applications, it is critical to know (1) which types of IHMI methods would be effective considering data availability and (2) what trade-offs between data availability and the IHMI model performance would result from applicable IHMI methods. However, generalized knowledge on these matters is still limited in civil engineering, which calls for research to optimize data collection efforts.

To bridge these knowledge gaps, we focus on identifying the reliable, safe, and efficient approaches to collecting human-related data via human-AI interfaces. The following research questions are proposed:
- What data are generated in civil engineering tasks?
- What types of data do AI in civil engineering need to process?
- What are efficient ways of processing structured data (e.g., spreadsheets, databases, and graphs) and unstructured data (e.g., text, imagery, sound, and sensor data)?
- What are efficient ways of annotating data that require domain knowledge to train AI?

*KG 4: The influence of IHMI on AEC industry and entrepreneurship*

KG4 lies in the area of the broader impact of IHMI on society. It is unknown what expected and unexpected influence will IHMI have on the AEC industry and entrepreneurship. The first aspect of this knowledge gap, which focuses on a macro point of view, is about the impact of AI on the industries related to the civil engineering domain, such as construction, transportation, and infrastructure. Existing literature has extensive discussions about identifying the impact of AI and automation on the economy. Although no definitive conclusion or well-accepted predictions exist, researchers agree that overall, AI will stimulate the economy's development, but its effect on different industries varies. Also, AI has a complex effect on the job market. For example, Furman and Seamans argue that AI may increase overall productivity and induce labor disruptions (Furman and Seamans 2019).

The second aspect, focusing on a micro point of view, is about the impact of IHMI on individuals, such as employers, employees, and educators. The unknown influence of IHMI on the economy and the industry will lead to unanswered questions about the influence on individuals' career related to civil engineering. Such questions include what jobs may disappear, what new jobs may emerge, what new tasks appear in existing jobs, and what new skills people need to develop. On top of existing works, correctly identifying or predicting the impact of IHMI may lead to well-designed policies to help the industries related to civil engineering better embrace the opportunities and mitigate the threats. Bridging this KG, however, is challenging because it requires novel data-driven socio-economic models. Such data are available only after the impact has occurred and may cause irretrievable harm to the economy, industry, and society.

Studies aiming at bridging this knowledge gap may focus on predicting the possible impact of IHMI on the AEC industry and thus identify ways to support a healthy labor market. Specifically, we propose the following research questions:
- How will human-AI integration change the job market? How can we deal with the reduced availability of low-skill jobs for humans that will result from increasingly capable machines?
- How will human-AI integration change the interest of stakeholders, the industry, and STEM education?

**CONCLUSIONS**

Integrating artificial intelligence's efficiency and high repeatability with human decision-makers' adaptability could advance timely and reliable civil engineering projects' decision-making. This paper shows the significance of exploring the value, method, and challenges of applying IHMI in civil engineering. Our systematic review of the literature and motivating cases identified the four knowledge gaps of IHMI for civil engineering. First, it is unknown what types of tasks in the civil engineering domain can be assisted by AI and to what extent. Second, the interface between human and AI in civil engineering-related tasks need more explicit and formal definition. Third, the barriers that impede collecting training data from humans and environments deserve systematic classification and prototyping of human-machine behavioral data collection. Lastly, it is unknown what expected and unexpected influence will IHMI have on society. A list of research questions to be answered to bridge these knowledge gaps. This paper will lay the foundation for identifying relevant studies in other fields to form a research roadmap to address the four knowledge gaps identified. The limitation of the paper - the lack of quantitative analysis - will be addressed by future exploratory works, including targeted literature review, a panel discussion, and surveys.